\documentclass{article}
\usepackage{spconf,amsmath,amssymb,graphicx,booktabs,url,hyperref,enumitem, multirow,subcaption,caption,lipsum,xcolor,array}

\hypersetup{hidelinks}
\newlength{\labw}   
\newlength{\slotw}  
\title{Time-Shifted Token Scheduling for Symbolic Music Generation}

\name{Ting-Kang Wang \qquad Chih-Pin Tan \qquad Yi-Hsuan Yang}
\address{Graduate Institute of Communication Engineering, National Taiwan University, Taiwan}

\begin{document}
\maketitle

\begin{abstract}


Symbolic music generation faces a fundamental trade-off between efficiency and quality. Fine-grained tokenizations achieve strong coherence but incur long sequences and high complexity, while compact tokenizations improve efficiency at the expense of intra-token dependencies. To address this, we adapt a \textbf{delay-based scheduling mechanism (DP)} that expands compound-like tokens across decoding steps, enabling autoregressive modeling of intra-token dependencies while preserving efficiency. Notably, DP is a lightweight strategy that introduces no additional parameters and can be seamlessly integrated into existing representations. Experiments on symbolic orchestral MIDI datasets show that our method improves all metrics over standard compound tokenizations and narrows the gap to fine-grained tokenizations.
\end{abstract}


\section{Introduction}
\label{sec:intro}

Symbolic music generation with Transformer-based language models has shown potential once MIDI sequences are tokenized into discrete events. A key challenge lies in designing effective tokenization schemes. REMI~\cite{huang2020pop, huang2024emotion, hung2021emopia} decompose a note into multiple attributes, such as pitch, duration, and velocity, producing flattened sequences that align well with language modeling. However, this leads to long token sequences, which substantially increase memory cost and inference latency.

To address this, token grouping strategies have been introduced. Compound Word Transformer~\cite{hsiao2021compound} demonstrated that grouping the attributes of a note into a single compound token can significantly shorten the sequence. In this framework, $K$ attribute embeddings are first combined (by summation or concatenation) into a single vector, then fed into the Transformer for sequence modeling. At the output stage, K parallel linear heads predict each attribute, thereby allowing all attributes to be decoded in parallel.

Similar multi-attribute representations have also appeared in the audio domain: MusicGen~\cite{copet2023simple} uses residual vector quantization (RVQ) to encode each audio frame into multiple coarse-to-fine layers, where each layer captures different levels of musical attributes. 

\begin{figure}
    \centering
    \includegraphics[width=\linewidth]{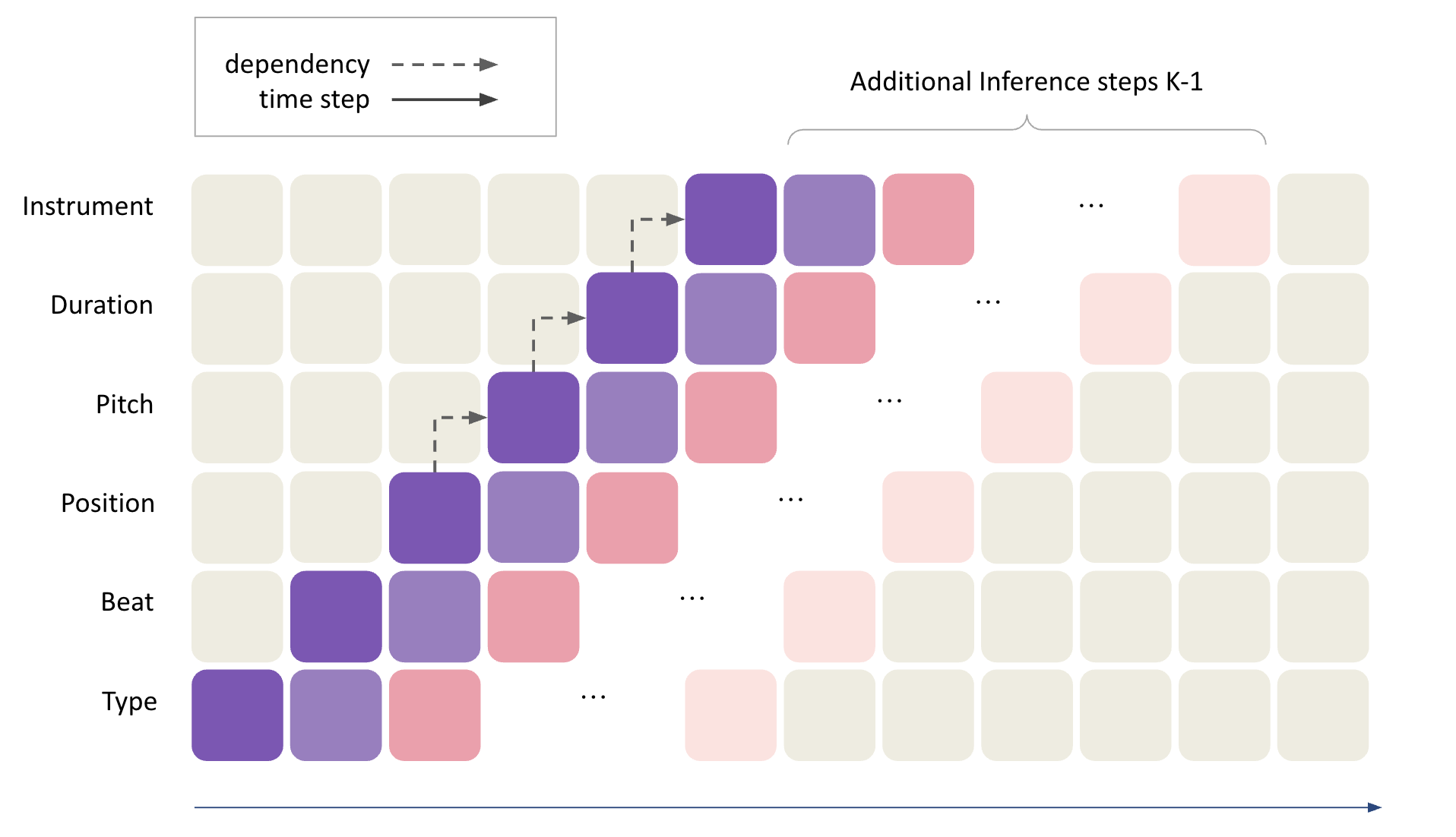}
    \caption{Visualization of delay-based interleaving mechanism on compound tokenization}
    \label{fig:delay-pattern}
\end{figure}

However, token grouping introduces a subtle trade-off. Since attributes are predicted in parallel from a single Transformer output, intra-token dependencies (e.g., pitch–duration correlation) may be ignored. 

To mitigate this issue, the Nested Music Transformer (NMT)~\cite{ryu2024nested} extends the standard Transformer with a sub-decoder module. Instead of predicting all attributes by parallel linear heads, the sub-decoder autoregressively unfolds each attribute from this hidden vector, using the outputs of previously predicted attributes as additional context. To strengthen this process, NMT introduces cross-attention mechanisms in its sub-decoder. The intra-token decoder performs cross-attention between the main decoder’s hidden state and the sequence of already generated attributes, ensuring that each new attribute prediction is conditioned on both the compound token context and its intra-token history. 
This allow NMT to model intra-token attributes correlations more effectively, achieving quality comparable to REMI while still benefiting from compact sequences. Yet, this improvement comes at a cost: the additional sub-decoders and cross-attention layers increase memory footprint and inference time, partly undermining the original motivation of token grouping.

This motivates us to ask: is there a lightweight approach that enhances intra-token dependency modeling without increasing model complexity? Inspired by MusicGen’s interleaving schedule for RVQ codebooks~\cite{copet2023simple}, we adapt a \textbf{delay-based interleaving mechanism (denoted as DP)} for symbolic music representation. This design allows the model to autoregressively capture dependencies between attributes while still preserving a single causal stream. Importantly, DP is not a new tokenization scheme, but rather a lightweight scheduling strategy that can be plugged into existing compound-like tokenizations, introducing no additional parameters and requiring only minimal changes to the data loader.

In experiments, we evaluate the scheduling mechanism by applying it to Multitrack Music Transformer (MMT)~\cite{dong2023multitrack}, a general multitrack extension of compound tokenization. We refer to this variant as MMT-DP and evaluate it on orchestral continuation tasks. Results show that applying DP narrow the quality gap to REMI representations while preserving lightweight inference efficiency. To ensure reproducibility and encourage adoption, we will open-source the implementation and offer an demo page\footnote{Demo Page: \url{https://tklovln.github.io/dptoken-demo/}} with our generated samples.

\begin{table}[t]
    \vspace{1.5em}
    \resizebox{\columnwidth}{!}{
        \centering
        \begin{tabular}{lcc}
            \toprule
            & \shortstack{Inference Speed \\ (Notes Per Second)} & \shortstack{Complexity  \\ (Big-O)} \\
            \midrule
            MMT~\cite{dong2023multitrack}         & 63.53 & $O(N^2)$ \\
            \textbf{MMT-DP (Ours)} & 62.47 & $O((N + (K-1))^2)$ \\
            NMT~\cite{ryu2024nested}          & 41.99 & $O(E^2) + O(NK)$ \\
            REMI+~\cite{von2022figaro}         & 20.42 & $O((NK)^2)$ \\
            \bottomrule
        \end{tabular}
    }
    \caption{Comparison of inference speed and complexity. 
    $N$ denotes the number of note events in a sequence, and $K$ is the number of sub-fields per token (e.g., type, beat, pitch, etc.).}
    \label{table:inference-speed}
\end{table}

\section{Related Work}
\label{sec:related work}

Language models, particularly Transformer-based architectures~\cite{vaswani2017attention}, have become dominant in natural language processing tasks~\cite{brown2020language}. Several studies have  adapted language modeling techniques to music. 
For example, \cite{ens2020mmm} proposed a MIDI-event-like representation that creates a time-ordered sequence for each track and concatenates multiple tracks into one sequence. 
In~\cite{zeng2021musicbert}, a permutation-invariant BERT-like language model was introduced for symbolic music understanding. In~\cite{liu2022symphony}, the author proposed a permutation-invariant architecture with a novel multitrack representation, 3-D positional embeddings, and a byte pair encoding scheme for music tokenization.
Building on these advances, symbolic music has been studied in various aspects. Some
prior work focused on unconditioned generation, including generating piano music \cite{huang2019music, huang2020pop}, guitar tabs \cite{chen2020automatic}, lead sheets \cite{wu2020jazz, wu2023compose} and multitrack music \cite{ens2020mmm, liu2022symphony, dong2023multitrack, ryu2024nested} from scratch. Others studied have explored controllable music
generation \cite{von2022figaro, shih2022theme}, song conver generation \cite{tan2024picogen, tan2024picogen2}, music style transfer \cite{wu2023musemorphose, le2024meteor}, polyphonic score infilling \cite{chang2021variable} and general-purpose pretraining for symbolic music
understanding \cite{wang2021musebert, zeng2021musicbert}.

\section{Proposed Method}
\label{sec:method}

\begin{figure}
  \centering
  \includegraphics[width=1\linewidth]{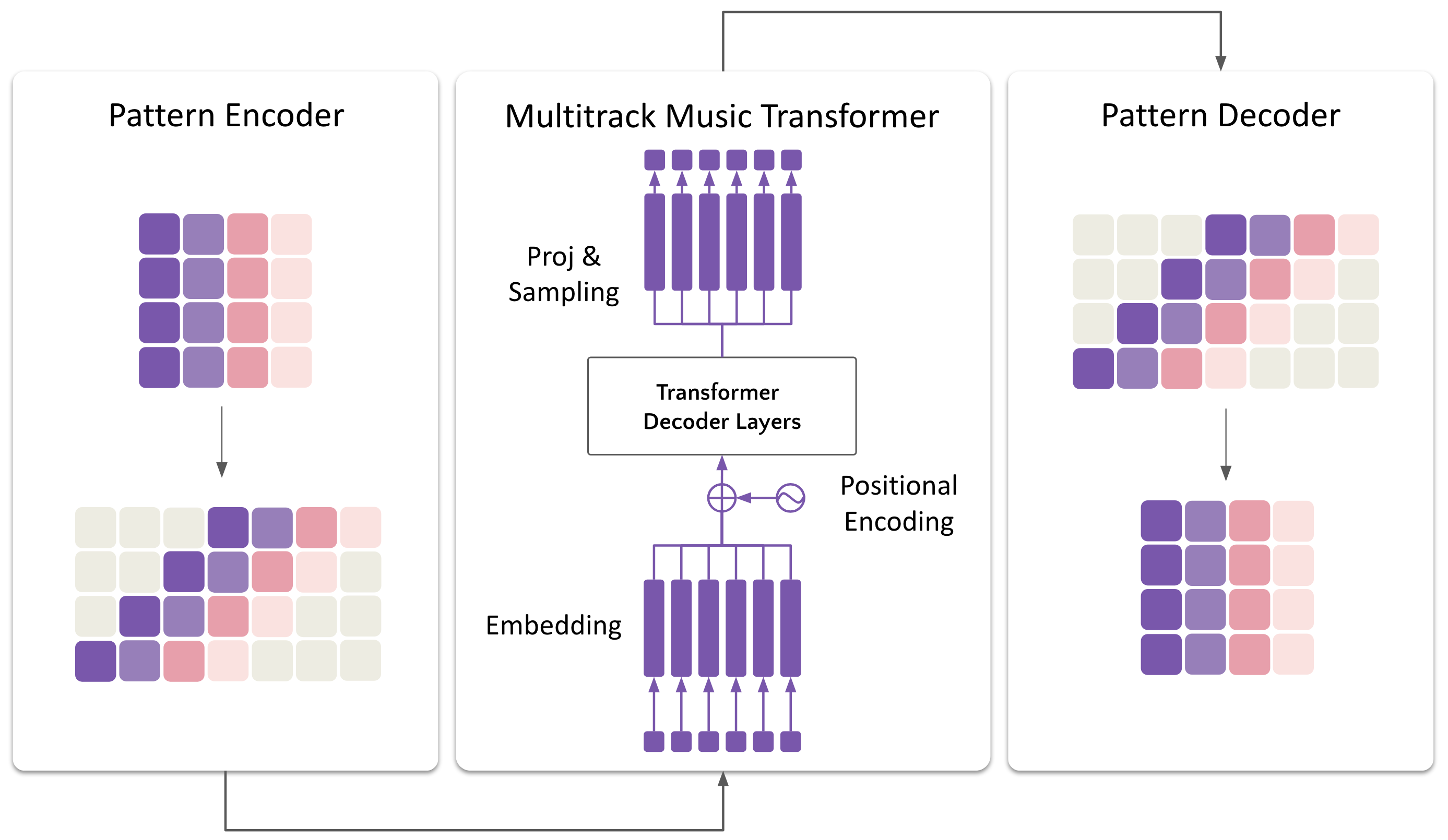}
  \caption{Overview of the system architecture. A delay-pattern encoder decoder framework is integrated into a multitrack music transformer.}
  \label{fig:model-architecture}
\end{figure}

\subsection{Token Representation}

Fig.~\ref{fig:delay-pattern} shows the music representation of our proposed delay scheduling, we adopt a compound representation with six discrete sub-fields per event:
\[
\resizebox{0.9\linewidth}{!}{
    \texttt{type},~\texttt{beat},~\texttt{position},~\texttt{pitch},~\texttt{duration},~\texttt{instrument}
}
\]
where onset $=$ \texttt{beat}$\cdot r +$\texttt{pos} at resolution $r=12$, following prior practice \cite{hsiao2021compound,huang2020pop}. In Figure~\ref{fig:delay-pattern}, horizontal axis corresponds to the autoregressive inference time steps of the transformer decoder, while the vertical axis denotes the sub-fields within each compound token. Dashed lines indicate potential dependencies among musical attributes across different sub-fields, illustrating how structural relations may extend beyond the strict autoregressive order. It can also be observed that this modification to the compound token only increases the token sequence by a constant term, namely $K{-}1$.

\subsection{Delay Scheduling Mechanism}
Given event $e_i=\{e_i^{(1)},\dots,e_i^{(6)}\}$, we assign fixed delays $\{\Delta_d\}$ and decode fields over adjacent steps:
\begin{align}
t &= i+\Delta_d, \\
p(e_i^{(d)} \mid \text{ctx}_{<t}) &= p\!\left(e_i^{(d)} \mid \{e_i^{(d')}\!:\!\Delta_{d'}\!<\!\Delta_d\}, ~\text{events }<i\right).
\end{align}
We set $\Delta_{\texttt{type}}{=}0$, then $\Delta_{\texttt{beat}}{=}1$, $\Delta_{\texttt{pos}}{=}2$, $\Delta_{\texttt{pitch}}{=}3$, $\Delta_{\texttt{duration}}{=}4$, $\Delta_{\texttt{instrument}}{=}5$. The joint factorization for one event becomes:
\begin{equation}
\small
p(e_i)=\prod_{d=1}^{6} p\!\left(e_i^{(d)} \,\middle|\, e_i^{(1{:}d-1)},~ e_{<i}\right).
\end{equation}

In addition, we explored different $\Delta_d$ settings and permutations of sub-field ordering. Ultimately, we found that using a uniform step delay with the current ordering provides the best empirical performance while preserving the natural causal structure of the representation.

\subsection{Architecture and Objective}
Following \cite{dong2023multitrack}, we use a decoder-only Transformer with intra-token embeddings summed into a compound embedding, plus absolute positional embeddings. A lightweight output heads map the hidden state to each sub-field’s vocabulary. We train with teacher forcing and the sum of cross-entropies over sub-fields and positions; inference uses top-$k$ sampling per field with monotonicity constraints on \texttt{type} and \texttt{beat}.%

\section{Experiments}
\label{sec:exp}

\subsection{Datasets}
\label{sec:data}
We conduct our experiments on \textbf{SymphonyNet}~\cite{liu2022symphony}, a dataset comprising 46,359 orchestral MIDI scores with an average duration of 256 seconds per piece, amounting to a total of 3,284 hours. The dataset features multitrack arrangements that span a wide range of time signatures and encompass both contemporary and classical orchestration styles.
For data augmentation we apply random pitch transpositions \(s \sim \mathcal{U}(-5, 6)\) (semitones) to improve robustness. Each dataset is split into training, validation and test sets with an 80\% / 10\% / 10\% ratio. For preprocessing and visualization, 
we use the \texttt{muspy}~\cite{dong2020muspy} and \texttt{pypianoroll}~\cite{dong2018pypianoroll} libraries.

\subsection{Models and Training}
\label{sec:model-config}
We train our models under the same configuration as MMT: 
8 layers, 8 attention heads, $d_{\text{model}}=512$, feed-forward dimension of 2048, 
dropout rate 0.1, and the Adam optimizer with an initial learning rate of $3\times 10^{-4}$, 
linear warmup followed by inverse square root decay. 
Training is conducted with a maximum sequence length of 1024, batch size 16, for 200k steps. 
For the NMT baseline, we instead adopt the settings reported in its original paper. 
We compare (a) MMT,(b) MMT with delay-pattern scheduling (denoted as MMT-DP), (c) REMI+\cite{von2022figaro}, a multitrack extension of REMI, and 
(d) NMT.

\section{Results}

\begin{figure}
    \centering
    \includegraphics[width=\linewidth]{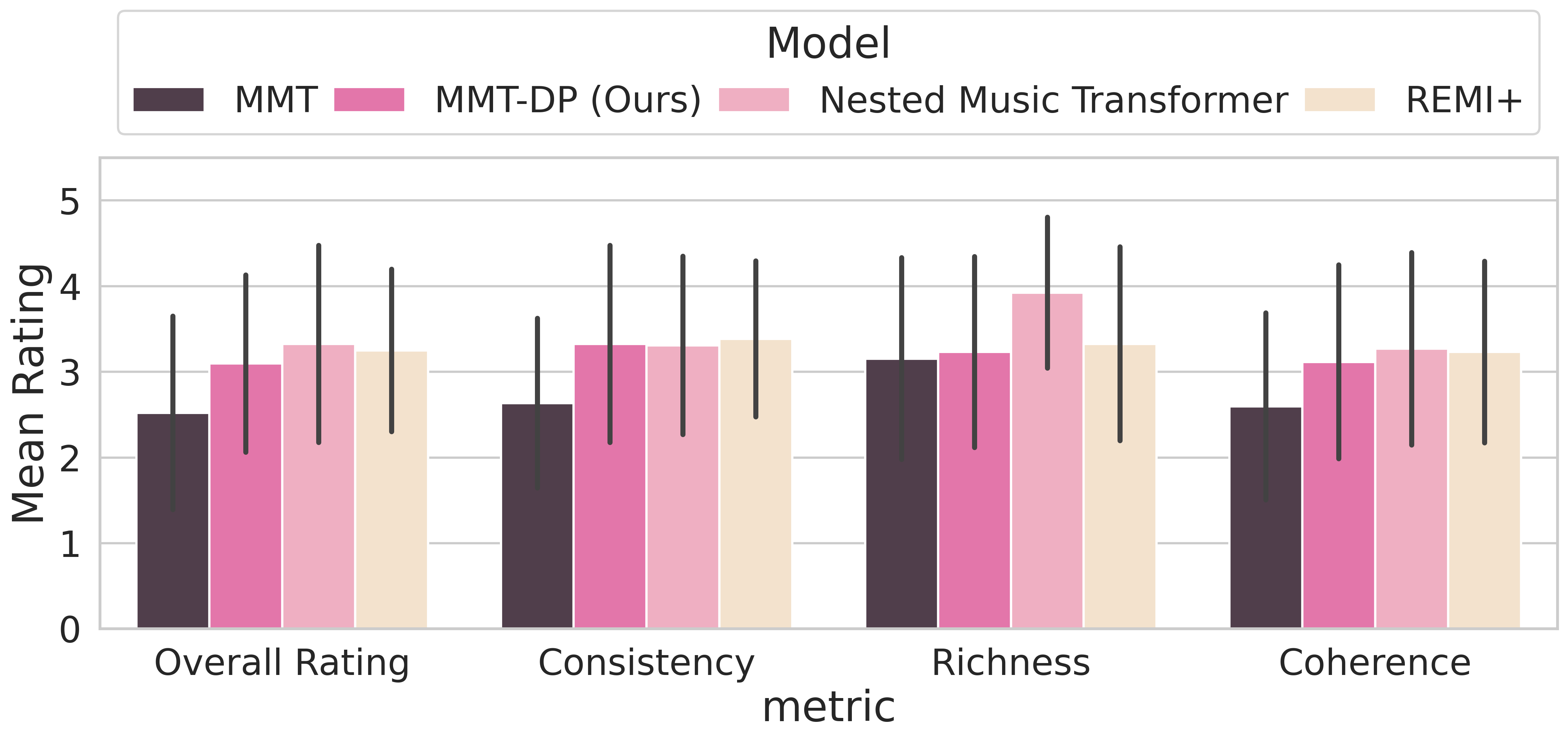}
    \caption{Orchestra generation MOS with 95\% CI}
    \label{fig:subjective-orchestra}
\end{figure}
\subsection{Subjective Listening Test} 
\label{sec: subject listening test}
\begin{figure*}[t]
  \centering
  \vspace{-1.5em}
  \begin{tabular*}{\linewidth}{@{\extracolsep{\fill}} m{0.2em} >{\centering\arraybackslash}m{0.45\linewidth} >{\centering\arraybackslash}m{0.45\linewidth}}
  & w/o DP & w/ DP \\
  \textbf{(a)} & \includegraphics[width=\linewidth, trim=0 500 0 0, clip]{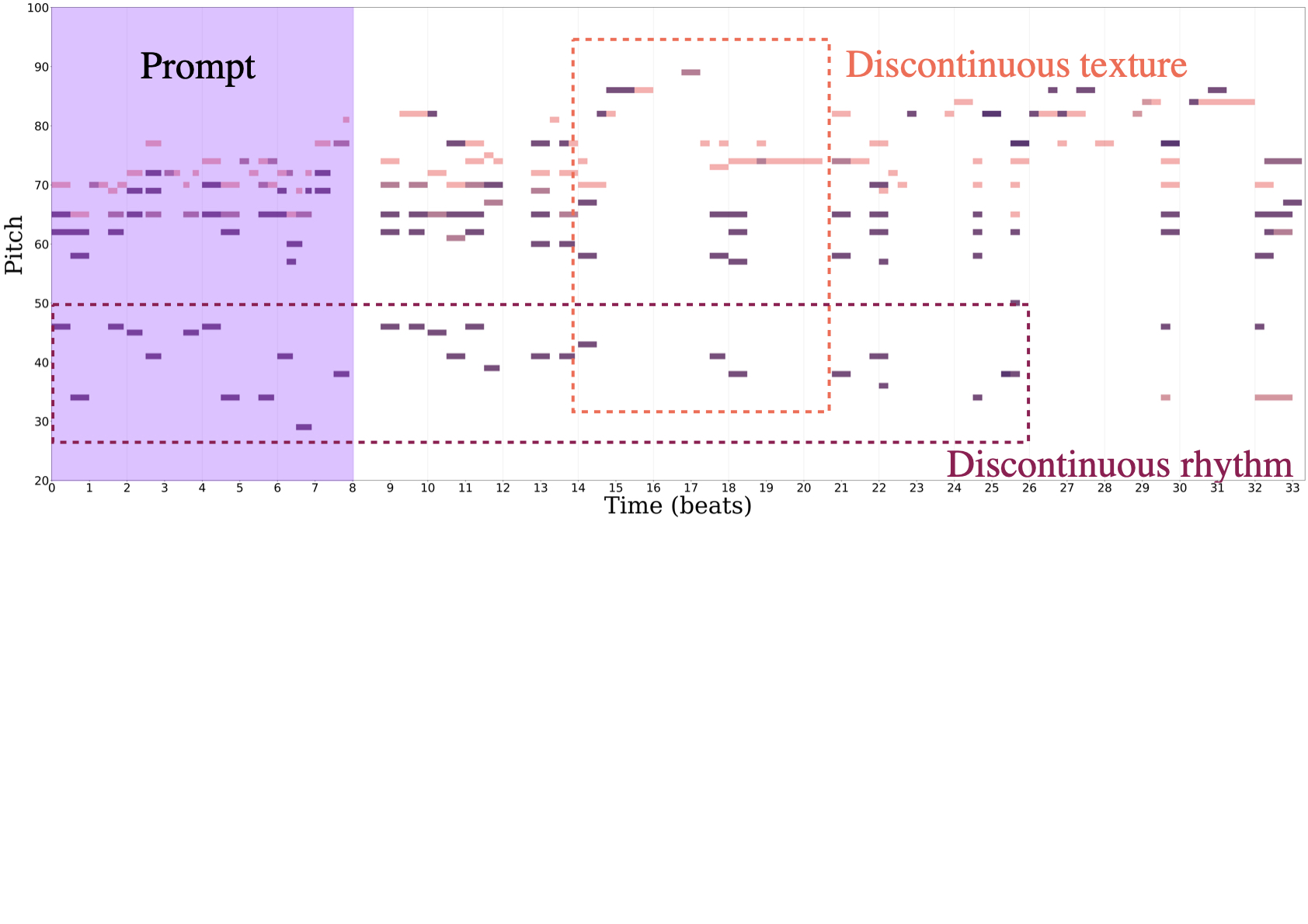} & \includegraphics[width=\linewidth, trim=0 500 0 0, clip]{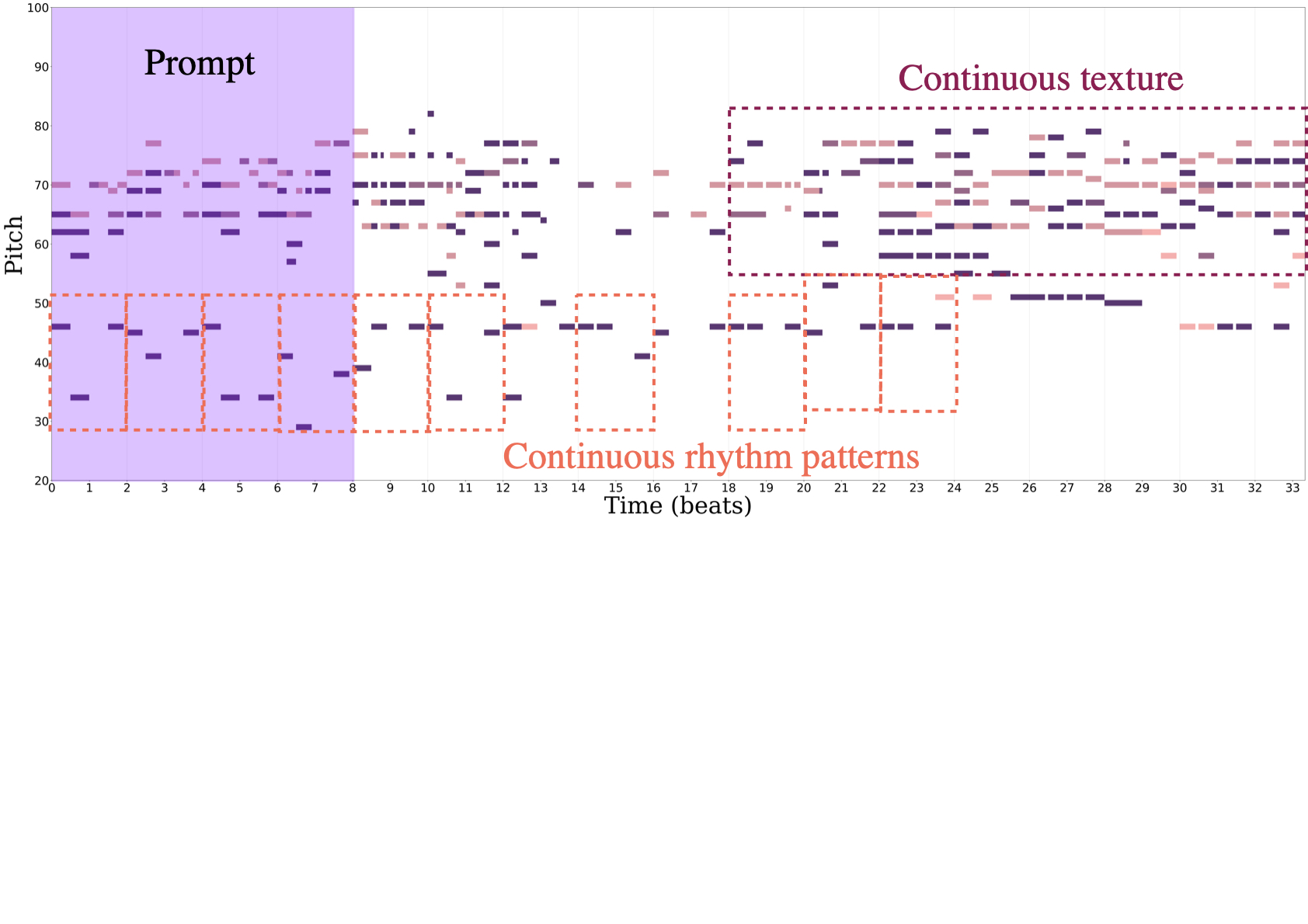} \\
  
  \textbf{(b)} & \includegraphics[width=\linewidth, trim=0 500 0 0, clip]{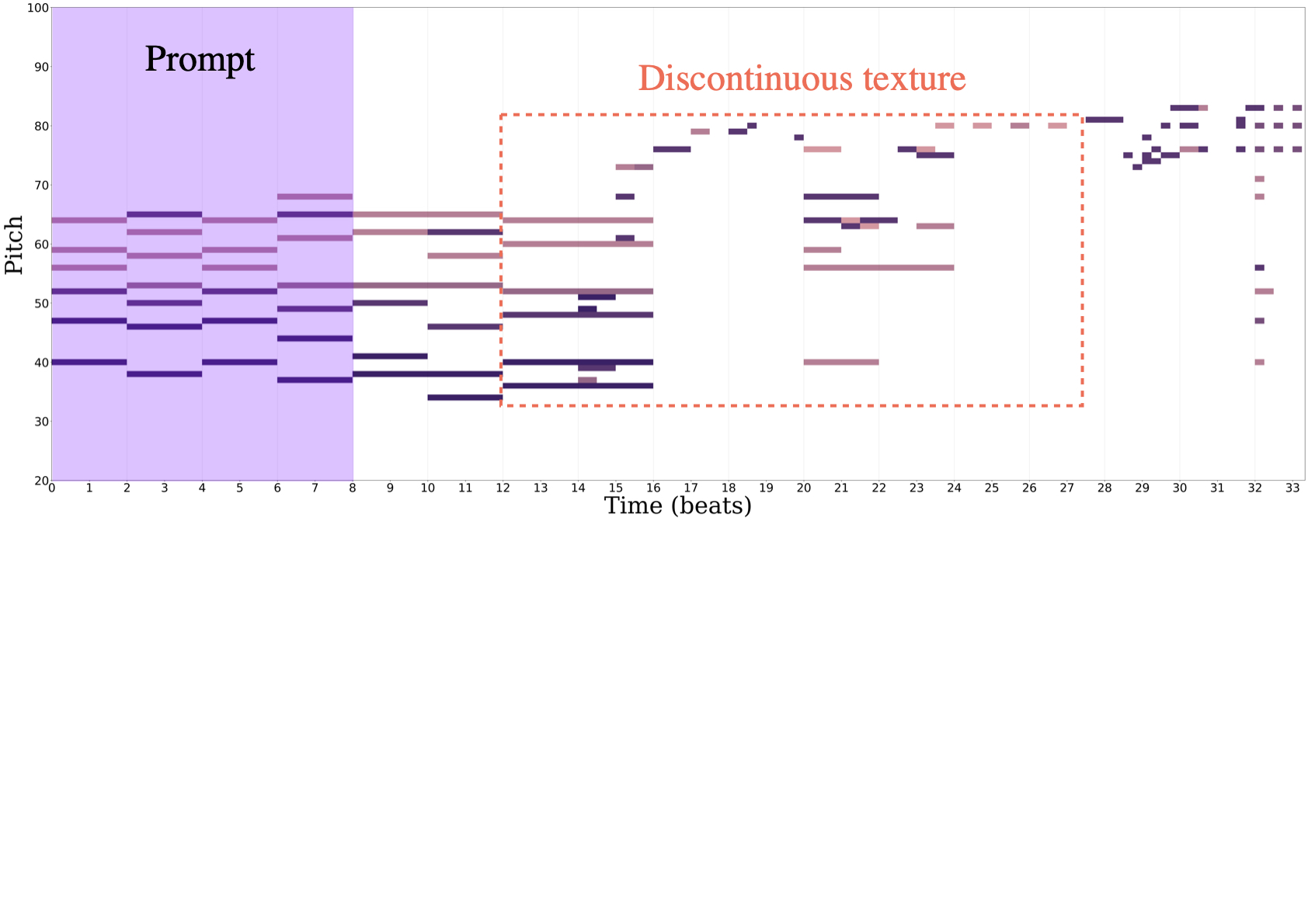} & \includegraphics[width=\linewidth, trim=0 500 0 0, clip]{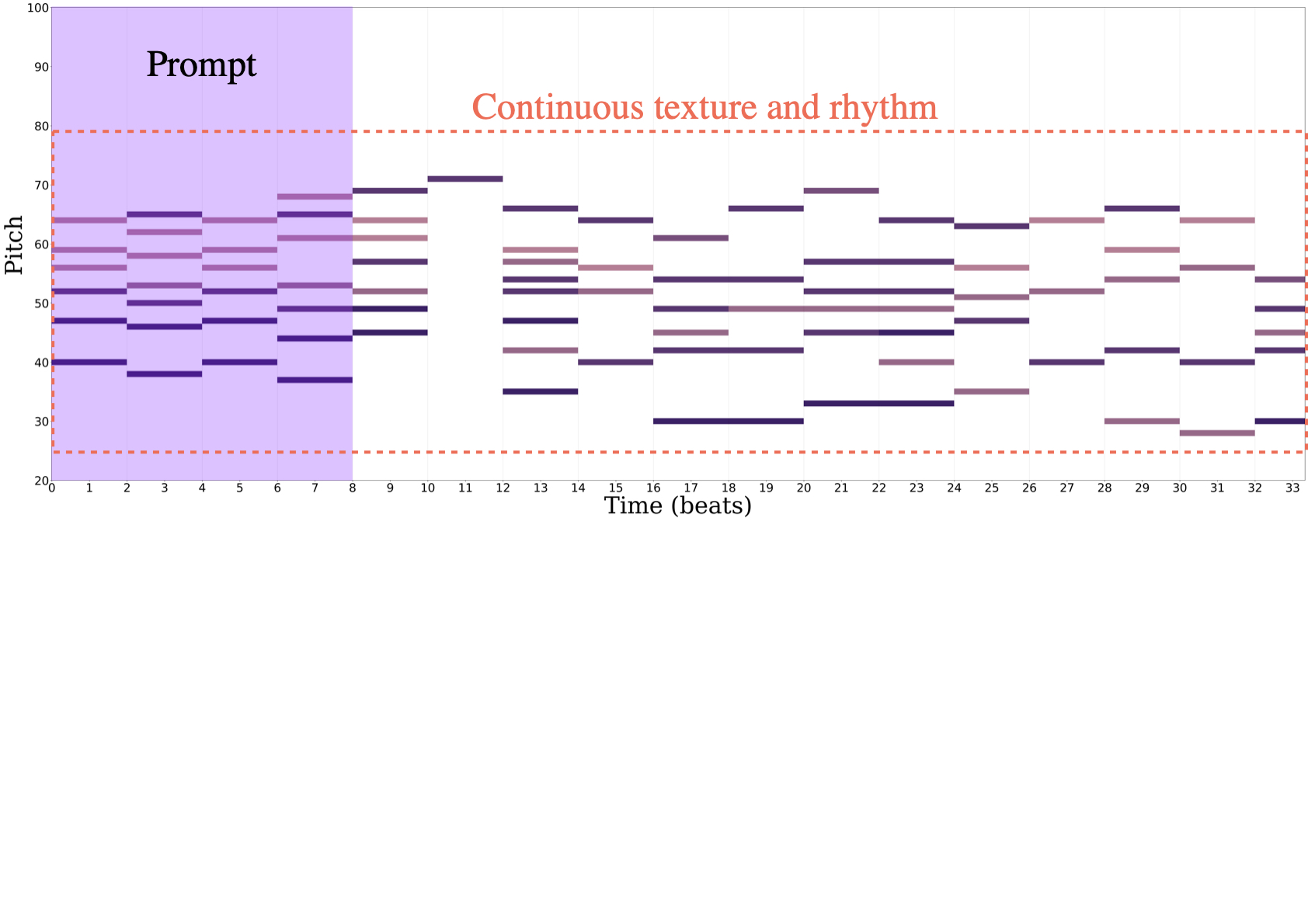} \\
  
  \textbf{(c)} & \includegraphics[width=\linewidth, trim=0 500 0 0, clip]{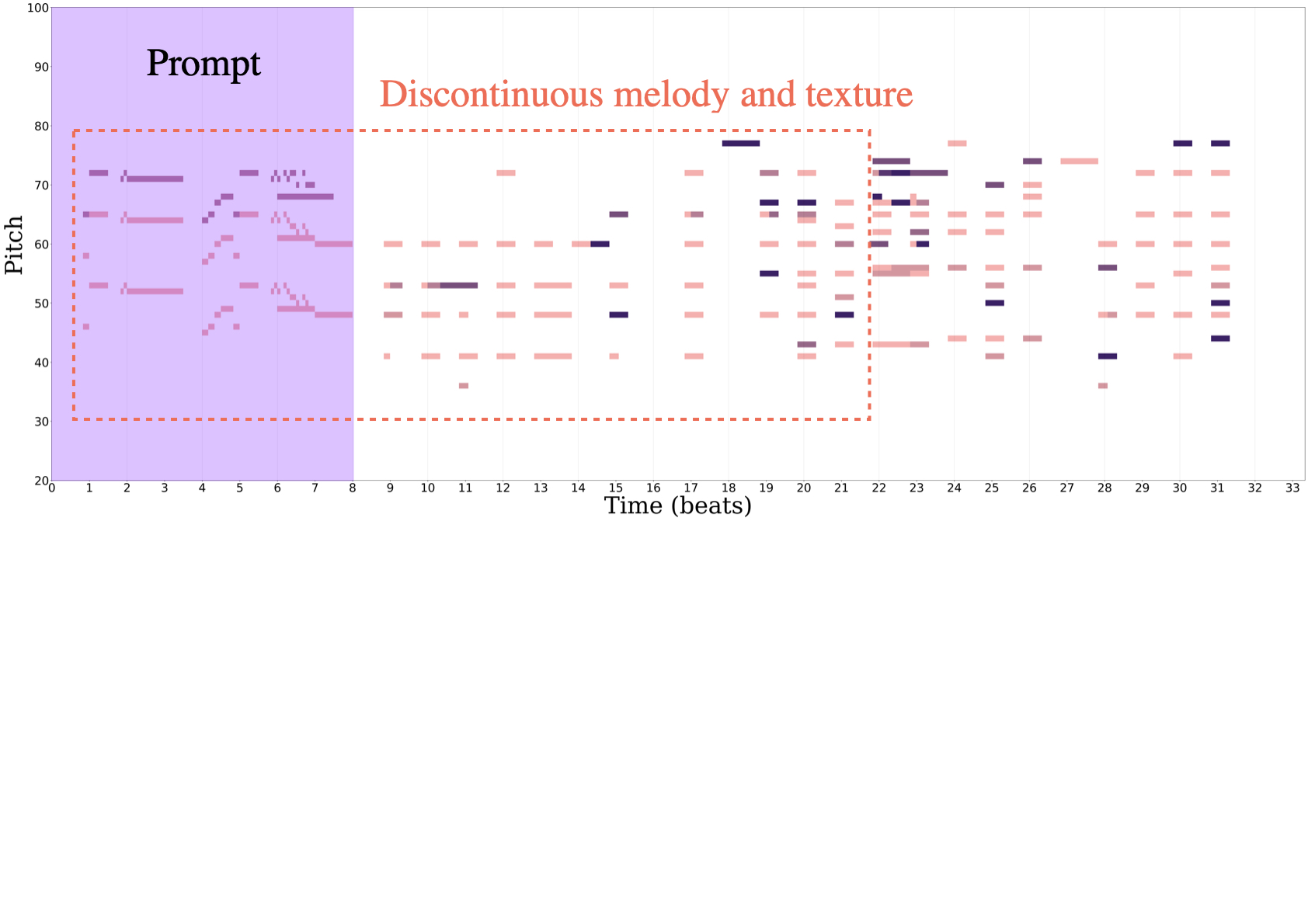} & \includegraphics[width=\linewidth, trim=0 500 0 0, clip]{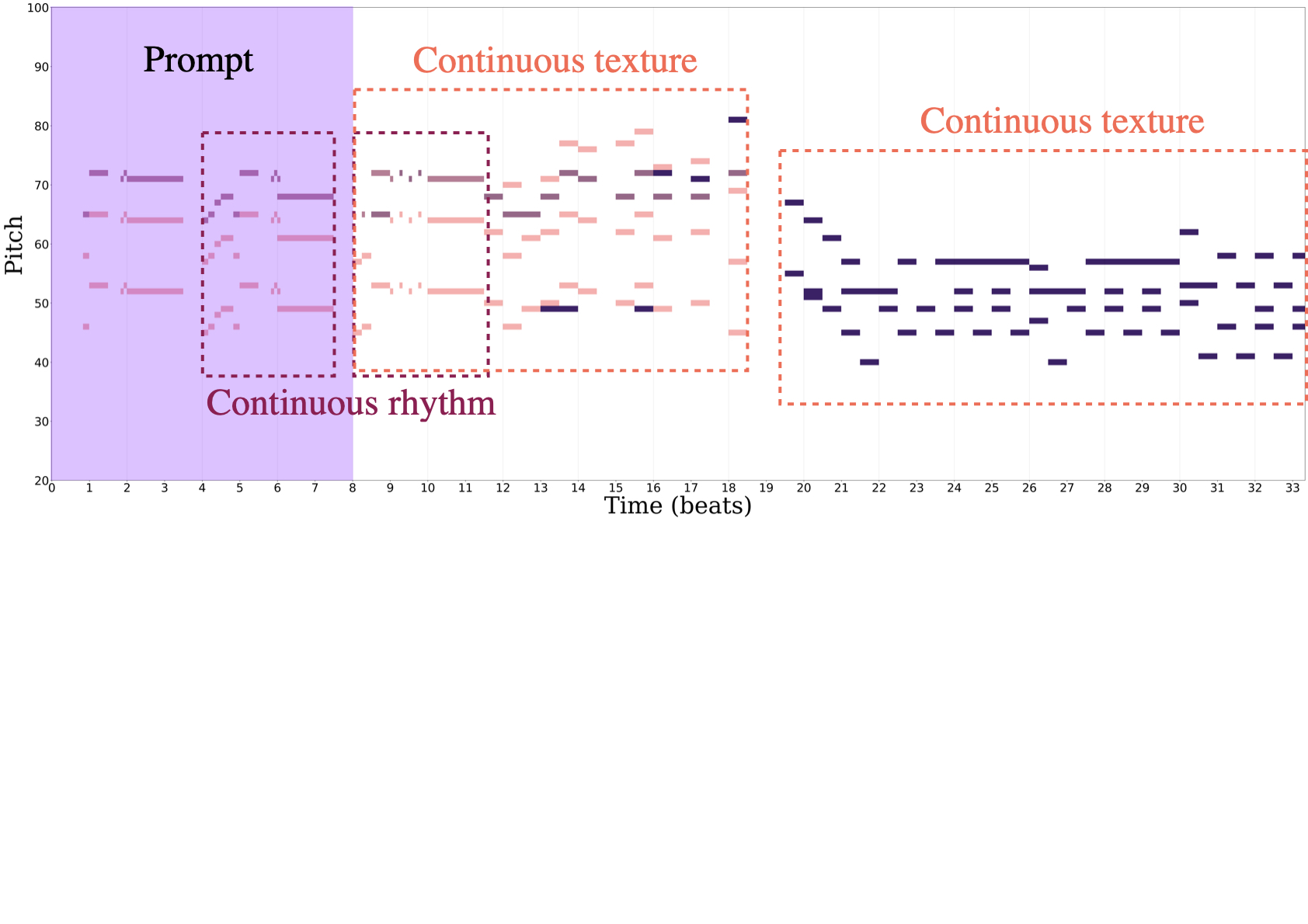} \\
  
\end{tabular*}
  \caption{Comparison of 3 test prompts, each row showing outputs from the same model w/o (left) and w/ (right) delay scheduling. The shaded region indicating the prompt, and colors indicate different instrument tracks.} 
  \label{fig:comparison}
\end{figure*}



To assess perceptual quality, we conducted a blind listening test with 26 participants (each playing at least one instrument). Among the participants, 16 had advanced musical training with over four years of experience, 5 had between one and three years of training, and 5 had less than one year of experience. We adopted a 2-bar continuation setup: for each sample, the first two measures of a MIDI prompt were provided to the model, which then generated the continuation until reaching either 1024 time steps or an end-of-sequence token. All MIDI outputs were rendered to audio using the \texttt{fluidsynth} library.

Each participant was presented with 2 randomly selected samples. For each sample, participants first heard the 2-bar prompt, followed by continuations from 4 models 
presented in randomized order. Ratings were collected on a 1–5 scale, where higher values indicate better quality, across four dimensions: (i) \textbf{Coherence}, reflecting the smoothness of transitions and phrasing; (ii) \textbf{Richness}, capturing the variety and interest of texture and harmony; (iii) \textbf{Consistency}, assessing the absence of compositional errors and overall unity; and (iv) \textbf{Overall Rating}, representing the general impression of the continuation.

Fig.~\ref{fig:subjective-orchestra} shows the mean opinion scores (MOS) of all models. We observe that applying delay scheduling leads to steady improvements over the MMT baseline across all metrics, especially in consistency, coherence, and overall listening quality. 
The results reach a level comparable to NMT and REMI+, confirming the strength of the proposed method. These findings highlight the benefit of applying delay scheduling to compound tokenization, which helps narrow the quality gap between compact compound tokens and fine-grained tokenization.


\subsection{Objective Evaluation} 
\begin{table}[t]
    \centering
    \resizebox{\linewidth}{!}{%
        \begin{tabular}{lccc}
            \toprule
            \textbf{Model} & \textbf{Pitch Class Entropy} & \textbf{Scale Consistency} & \textbf{Groove Consistency} \\
            \midrule
                Ground truth & 2.70 ($\pm$0.39) & 0.92 ($\pm$0.08) & 0.90 ($\pm$0.07) \\
                MMT~\cite{dong2023multitrack} & 2.42 ($\pm$0.46) & 0.96 ($\pm$0.05) & 0.90 ($\pm$0.07) \\
                NMT~\cite{ryu2024nested} & 2.74 ($\pm$0.43) & 0.92 ($\pm$0.07) & 0.99 ($\pm$0.00) \\
                REMI+~\cite{von2022figaro} & 2.64 ($\pm$0.46) & 0.92 ($\pm$0.07) & 0.88 ($\pm$0.08) \\
                MMT-DP (Ours) & 2.53 ($\pm$0.46) & 0.95 ($\pm$0.06) & 0.93 ($\pm$0.05) \\
            \bottomrule
        \end{tabular}%
    }
    \caption{Objective evaluation results on orchestra generation.}
    \label{table:objective-combined}
\end{table}

In addition to the subjective listening test, we follow \cite{dong2023multitrack,ryu2024nested} and measure pitch class entropy, scale consistency, and groove consistency to evaluate the performance of the proposed model. For these metrics, values closer to the ground truth are considered better. As shown in Table~\ref{table:objective-combined}, our generated samples achieve scores that are consistently close to the ground truth across all three dimensions, suggesting that the model is able to produce musically plausible outputs.  

\subsection{Case Study on Multitrack Generation}
To demonstrate the effectiveness of our approach, we present a case study on a 2-bar continuation task, using the same configuration described in Section~\ref{sec: subject listening test}. 
Figure~\ref{fig:comparison} illustrates examples of 2-bar continuation, comparing outputs generated with and without delay-pattern scheduling (DP). 

In case (a), sample without DP quickly diverges from the prompt, exhibiting discontinuous texture and fragmented rhythm patterns. With DP, however, both rhythmic and textural structures remain consistent. In case (b), the baseline maintains continuous texture and rhythm at first, but later produces abrupt gaps and unstable patterns. The DP variation model mitigates this by producing rhythmically stable sequences. In case (c), we observe that the melody line is scattered across the continuation and fails to align with the prompt’s texture and rhythmic pattern, leading to incoherent pieces. In contrast, samples with DP preserves rhythm patterns at the beginning and maintain consistent textures by multiple measures, producing a more coherent and musically plausible continuation. 

Beyond improving stability and coherence, the continuations also demonstrate creativity through novel chord progressions and melodic development, showing that the model is not merely imitative but musically expressive.

\subsection{Inference-time Complexity}
In addition to quality, a key advantage of our approach is its lightweight nature at inference time. We evaluated efficiency by generating sequences up to 1024 steps or until an EOS token on a single RTX~3090 GPU with default decoding settings in Section~\ref{sec:model-config}. For each piece, we computed notes-per-second (NPS) and averaged across the test set (see Section~\ref{sec:data}). As shown in Table~\ref{table:inference-speed}, MMT reaches 63.53 NPS, while MMT-DP achieves 62.47 (only 1.7\% slower). In contrast, NMT runs at 41.99 NPS and REMI+ at 20.42. These results confirm that delay scheduling adds virtually no overhead, with MMT-DP over 50\% faster than NMT and nearly $3\times$ faster than REMI+, making it suitable for real-time applications.

\section{Conclusion}
We demonstrate that the delay-based scheduling mechanism, previously explored in the audio domain, can also be applied symbolic music representation. It strengthens intra-token dependency modeling without adding complexity or inference overhead, and achieves promising results on orchestra generation. For future work, we plan to further explore how different delay orders influence music generation models, as well as validate the approach across broader musical styles and specific instrument corpora such as piano.

\clearpage
\small
\bibliographystyle{IEEEbib}
\bibliography{refs.bib}

\end{document}